%% file: main.tex
\documentclass[10pt,twocolumn,letterpaper]{article}

\usepackage[pagenumbers]{cvpr}      

\input{preamble}

\usepackage{amsmath}
\usepackage{amssymb}
\usepackage{graphicx}
\usepackage{bm}
\usepackage{geometry}
\usepackage[table]{xcolor}
\usepackage{tikz}
\usetikzlibrary{arrows.meta, positioning}
\usepackage{multirow}

\definecolor{mygray}{gray}{0.9}

\DeclareMathAlphabet\mathbfcal{OMS}{cmsy}{b}{n}

\def\0{{\bf 0}}
\def\1{{\bf 1}}

\def\bI{{\bf I}}

\def\bM{{\bf M}}

\def\bx{{\bf x}}

\def\bx{{\bf x}}

\usepackage{ntheorem}

\newtheorem*{*thm}{Theorem}

\newtheorem*{*lemma}{Lemma}

\definecolor{cvprblue}{rgb}{0.21,0.49,0.74}
\usepackage[pagebackref,breaklinks,colorlinks,allcolors=cvprblue]{hyperref}

\def\paperID{31170} 
\def\confName{CVPR}
\def\confYear{2026}

\title{Efficient Matrix Implementation for Rotary Position Embedding}

\author{
  Chen Minqi$^{1}$\thanks{Corresponding author: \texttt{chenminqi@huawei.com}} \quad
  Zhongqi Yue$^{2}$ \quad
  Shihao Zhang$^{1}$ \quad
  Yun Xu$^{1}$ \quad\\
  Peng Wu$^{1}$ \quad
  Kaixiang Xu$^{1}$ \quad
  Zeyi Huang$^{1}$ \quad
  Hanwang Zhang$^{2}$ \\[0.5em]
  $^{1}$Huawei Technologies \quad
  $^{2}$Nanyang Technological University
}

\begin{document}

\maketitle
\input{sec/0_abstract}

\input{sec/1_intro}

\input{sec/2_related_work}
\input{sec/3_prelim}

\input{sec/4_method}
\input{sec/5_exp}
\input{sec/6_conclusion}
\input{sec/X_suppl}
{
    \small
    \bibliographystyle{ieeenat_fullname}
    \bibliography{main}
}


\end{document}

%% file: preamble.tex
\newcommand{\nd}[1]{{\color{Maroon}#1}}

\usepackage[ruled,vlined]{algorithm2e}



%% file: sec/0_abstract.tex
\begin{abstract}
Rotary Position Embedding (RoPE) has become a core component of modern Transformer architectures across language, vision, and 3D domains. However, existing implementations rely on vector-level {split} and {merge} operations that introduce non-negligible computational overhead, often overlooked in attention optimization. The problem is further amplified in multi-dimensional settings (\eg, 2D and 3D RoPE), where additional vector operations and uneven feature partitions degrade hardware utilization.  
To overcome these limitations, we propose \textbf{RoME} (\underline{Ro}tary \underline{M}atrix position \underline{E}mbedding), a mathematically equivalent yet computationally efficient reformulation of RoPE that replaces vector operations with unified matrix transformations. RoME eliminates dimension-specific operations, simplifies implementation, and enables fused parallel execution across Cube and Vector units on modern NPUs. Experiments show that RoME delivers substantial acceleration at both the operator and full-model levels. The implementation is available \href{https://gitcode.com/cann/ops-transformer/blob/master/experimental/posembedding/rope_matrix/README.md}{here}.

\end{abstract}

%% file: sec/1_intro.tex
\section{Introduction}
\label{sec:1}

Transformers have become the unified backbone across language, vision, and 3D domains, relying on positional embeddings to encode sequence order~\cite{vaswani2023attentionneed,dosovitskiy2021imageworth16x16words,liu2022petrpositionembeddingtransformation}. Traditional positional embeddings are tied to absolute positions~\cite{Radford2018ImprovingLU,devlin2019bertpretrainingdeepbidirectional,wang2020positionembeddingslearnempirical}, causing mismatches when sequences extend beyond the training length or undergo shifts. Rotary Position Embedding (RoPE)~\cite{Su2021RoFormerET} addresses this by encoding positions through rotations that depend only on relative offsets, making attention weights invariant to global shifts and enabling better generalization to unseen sequence lengths or spatial configurations.

\input{fig/fig1}

To understand RoPE, we consider its formulation for a 1-D sequence. Let $\mathbf{q}_a, \mathbf{k}_b \in \mathbb{R}^d$ denote the query and key features of length $d$ at sequence positions $a$ and $b$, respectively.
RoPE applies a position-dependent transformation $\text{RoPE}(\cdot)$ to each feature such that the resulting attention weight, $\text{RoPE}(\mathbf{q}_a, a)^{\!\top} \text{RoPE}(\mathbf{k}_b, b)$, depends only on the relative offset $(b - a)$ rather than their absolute positions. As illustrated in Figure~\ref{fig:1a}, the transformation consists of three steps: (1) \textbf{split} the input into groups of 2D vectors (\eg, by interleaving dimensions), (2) \textbf{rotate} each vector by an angle determined by its absolute position, and (3) \textbf{merge} all rotated vectors back to the original layout. After transformation, attention is computed in the standard way, \eg, using FlashAttention~\cite{dao2023flashattention2}.  
To extend RoPE to $n$-dimensional sequence (\eg, $n{=}2$ for images, $n{=}3$ for videos), one introduces an additional set of \nd{split} and \nd{merge} operations, color-coded in red to distinguish them from those within $\text{RoPE}$. Specifically, the input is first \nd{split} along the feature dimension into $n$ parts, $\text{RoPE}$ is applied independently to each part, and the outputs are then \nd{merged} back into the original structure~\cite{chu2024visionllamaunifiedllamabackbone,ma2025stepvideot2vtechnicalreportpractice}, as shown in Figure~\ref{fig:1b}.

\input{tables/tab1}

While the rotation in RoPE can be efficiently implemented using matrix operations, the accompanying split and merge steps---implemented as vector operations---introduce non-negligible computational overhead. This is often overlooked in prior works that primarily optimize the attention kernel itself. As shown in Table~\ref{table:MFU}, our profiling of a video generation model reveals that vector operations account for 22.5\% of total computation, with RoPE alone contributing 44\%. The inefficiency arises from three factors: (1) \textbf{split} operations are hardware-unfriendly (\eg, interleaving is suboptimal on Ascend devices); (2) higher-dimensional RoPE ($n>1$) introduces additional \nd{split} and \nd{merge} operations; and (3) for $n{=}3$, feature dimensions are typically uneven after \nd{splitting} (\eg, $d{=}128$ becomes 44, 44, and 40), further degrading hardware utilization (Section~\ref{sec:5.3}).

To address these inefficiencies, we propose a mathematically equivalent yet computationally optimized formulation, termed \underline{Ro}tary \underline{M}atrix position \underline{E}mbedding (\textbf{RoME}), which replaces the vector-level {split} and {merge} operations in RoPE with efficient matrix operations. We show that various RoPE implementations, differing in their splitting or interleaving strategies, can be unified under this matrix formulation. Furthermore, the same formulation naturally generalizes to 2D and 3D sequences, eliminating the need for dimension-specific \nd{split} and \nd{merge} operations. Our main contributions are summarized as follows:
\begin{itemize}
    \item We propose \textbf{RoME}, a matrix-based reformulation of RoPE that is mathematically equivalent yet computationally efficient. To the best of our knowledge, this is the first work to analyze RoPE at the operator level and study its interaction with hardware efficiency.
    \item For further acceleration, we implement a fused operator design and parallelize computations across the Cube and Vector units on modern NPUs.
    \item With these optimizations, RoME achieves up to $3\times$ speedup on origin RoPE implement, $1.48\times$ on fused operators, and an overall $5\%$ improvement in both training and inference throughput.
    \item RoME also greatly simplifies implementation, reducing the codebase for RoPE variants from 14k lines to about 100 lines.
\end{itemize}

%% file: fig/fig1.tex
\begin{figure}
    \centering
    \begin{subfigure}[t]{\linewidth}
         \includegraphics[width=\textwidth]{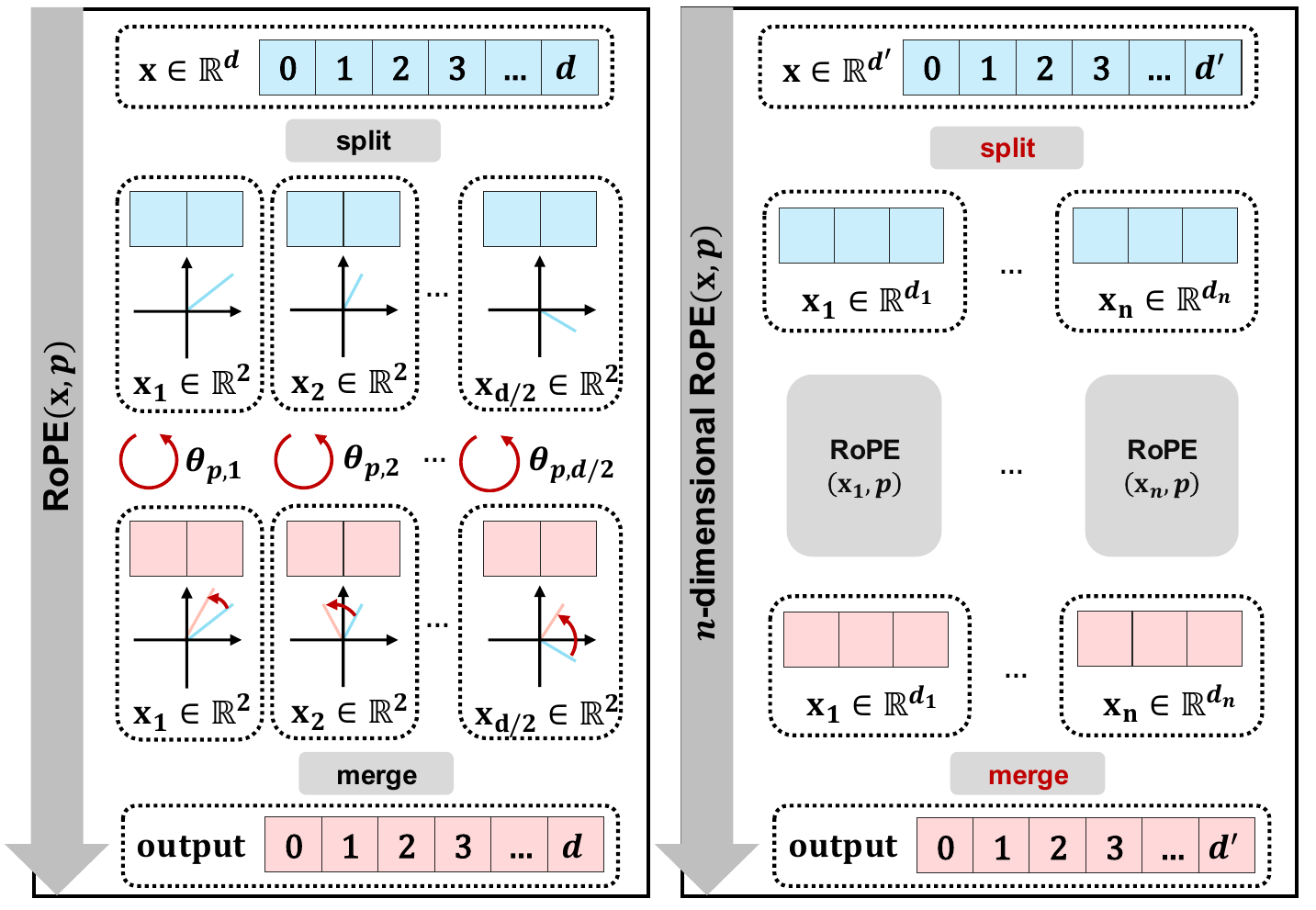}
         \phantomcaption
         \label{fig:1a}
    \end{subfigure}
    \begin{subfigure}[t]{0.\linewidth}
         \includegraphics[width=0\textwidth]{example-image-b}
         \phantomcaption
         \label{fig:1b}
    \end{subfigure}
    \vspace{-10mm}
    \caption{(a) Steps to compute $\text{RoPE}(\mathbf{x},p)$ given an input feature $\mathbf{x}\in\mathcal{R}^d$ and its absolute position $p$ for a 1-D sequence; (b) Steps for an $n$-D sequence, which involves an extra pair of \nd{split} and \nd{merge}. Note that $\sum_{i=1}^n d_i = d' $. }
    \label{fig:1}
    \vspace{-4mm}
\end{figure}

%% file: tables/tab1.tex
\begin{table}[t]
    \centering
    \caption{A performance profiling analysis based on Flux~\cite{labs2025flux} inference on Ascend 910B3. RoPE-related computations contributes $44\%$ of the total vector operator cost, making it a primary performance limiter.}
    \resizebox{\columnwidth}{!}{ 
    \begin{tikzpicture}[
      every node/.style={inner sep=0pt, outer sep=0pt, align=center},
      node distance=1.8cm and 0cm
    ]
    
    \node (main) {
    \begin{minipage}{0.46\columnwidth}
    \centering
    \begin{tabular}{c c c}
    \toprule
    Operators & s & \% \\
    \midrule
    FA/FAG & 0.348 & 34.8\%  \\
    Cube & 0.452 & 42.6\% \\
    \rowcolor{gray!20}
    \textbf{Vector} & \textbf{0.238} & \textbf{22.5\%} \\[-0.4em]
    \phantom{Vector} & & \\[-0.4em]
    Speed Gap & 0.001 & 0.1\% \\
    Overall & 1.059 & 100\%  \\
    \bottomrule
    \end{tabular}
    \end{minipage}
    };
    
    \node[right=2.2cm of main] (sub) {
    \resizebox{0.33\columnwidth}{!}{
    \begin{minipage}{0.40\columnwidth}
    \centering
    \begin{tabular}{c c c}
    \toprule
    Type & s & \% \\
    \midrule
    \textbf{RoPE} & \textbf{0.11} & \textbf{44\%} \\
    GeLU & 0.026 & 11\% \\
    LN & 0.022 & 9\% \\
    Trans& 0.021 & 8\% \\
    Concat & 0.021 & 8\% \\
    Mul & 0.13 & 6\% \\
    \multicolumn{3}{c}{......}\\
    \bottomrule
    \end{tabular}
    \end{minipage}
    }
    };

    \draw[-{Latex[length=2.5mm]}, thick, gray!70]
    ([xshift=14mm, yshift=-1ex]main.east) -- ([yshift=10ex]sub.west);

    \draw[-{Latex[length=2.5mm]}, thick, gray!70]
    ([xshift=14mm, yshift=-1.5ex]main.east) -- ([yshift=-10ex]sub.west);
    \end{tikzpicture}
    }
    \label{table:MFU}
    \vspace{-5mm}
\end{table}

%% file: sec/2_related_work
\section{Related Work}
\paragraph{Rotary Position Embedding(RoPE).}
RoPE is first proposed to for natural language Processing~\cite{Su2021RoFormerET} to impove ability for machine translation tasks. As times go on, researches find it's generalization ability and it quickly becomes one of the industry standards for LLM networks designing ~\cite{grattafiori2024llama3herdmodels,yang2025qwen3technicalreport,Liu2024}. 
Due to the explosive growth in the multimodal domain with transformer blocks, technique like RoPE are introduce step by step into these fields. 
A new version named RoPE2D is introduced into vision transformer~\cite{heo2024rotarypositionembeddingvision}, which inspire researches on generation domain with multimodal. The sequel of stable diffusion3 named FLux~\cite{labs2025flux} first introduce RoPE to generation domain and achieve promising result. 
After that, most image and video generation task(QwenImage~\cite{wu2025qwenimagetechnicalreport}, HunYuanVideo~\cite{kong2025hunyuanvideosystematicframeworklarge} and Wan~\cite{wan2025}) take RoPE as a default selection for architecture design. 
\paragraph{Transformer Acceleration.}
Since Transformer block becomes one of most import component for model designing, massive works are designed for it's acceleration. Such as quantization~\cite{li2024svdquant,xiao2023smoothquant,sun2024flatquant}, sparse-attention~\cite{zhao2025paroattentionpatternawarereorderingefficient,lou2024sparserfastermoreefficient,Liu2024} and hardware affinity designing algorithms~\cite{dao2022flashattention,dao2023flashattention2,zhang2024sageattention,zhang2024sageattention2,zhang2025sageattention3}. 
The success of FlashAttention~\cite{dao2023flashattention2} indicating that hardware affinity designing is critical for transformer acceleration. 
Thus, by carefully analysis most famous net with transformer, we believe RoPE acceleration is the most import task beside with FlashAttention acceleration.

%% file: sec/3_prelim.tex
\section{Preliminaries}

\paragraph{Rotary Position Embedding.}
Rotary Position Embedding (RoPE)~\cite{Su2021RoFormerET} divides the $d$-dimensional feature space into $d/2$ two-dimensional subspaces, and encodes positional information by applying a rotation within each subspace.
Formally, for an input feature vector $\bx \in \mathbb{R}^d$ at position $p$, RoPE performs
\begin{align}
\label{eq:rope}
\text{RoPE}(\bx, p) = R(\boldsymbol{\theta}_p)\, \bx, 
\end{align}
where $\boldsymbol{\theta}_p = [\theta_{p,1}, \ldots, \theta_{p,d/2}]$ and $R(\boldsymbol{\theta}_p)$ is a block-diagonal rotation matrix defined as:


\begin{align}
\label{eq:rope_details}
\begingroup
\setlength{\arraycolsep}{3pt} 
\renewcommand{\arraystretch}{0.9} 
\begin{bmatrix}
\cos\theta_{p,1} & -\sin\theta_{p,1} & \cdots & 0 & 0\\
\sin\theta_{p,1} &  \cos\theta_{p,1} & \cdots & 0 & 0\\
\vdots & \vdots &  \ddots & \vdots & \vdots\\
0 & 0 &  \cdots & \cos\theta_{p,d/2} & -\sin\theta_{p,d/2}\\
0 & 0 &  \cdots & \sin\theta_{p,d/2} & \cos\theta_{p,d/2}\\
\end{bmatrix}
\endgroup
\end{align}
where $\theta_{p, i} = p \omega_i$, and $\omega_i$ denotes the frequency associated with the $i$-th two-dimensional subspace. In practice, $\omega_i$ is typically set to $10000^{-2(i-1)/d}$.
Intuitively, each pair of adjacent dimensions $(x_{2i}, x_{2i+1})$ is rotated by a distinct angle proportional to its position $p$, enabling position-dependent phase modulation while preserving the vector norm and cosine similarity.

\paragraph{Implementation.} In practice, the RoPE operation is applied to a sequence of embeddings 
$\mathbf{X} = [\mathbf{x}_1, \ldots, \mathbf{x}_s] \in \mathbb{R}^{s \times d}$, 
where $s$ denotes the sequence length. The standard implementation exploits 
the sparsity of $R(\boldsymbol{\theta}_p)$ and converts the matrix multiplication 
in Equation~\ref{eq:rope} into efficient element-wise vector operations 
using tensor slicing and concatenation:
\begin{equation}
\begin{aligned}
&\mathbf{X}_1, \mathbf{X}_2 = \texttt{split}(\mathbf{X}, -1, 2), \\
&\mathbf{X}_{\text{new}} = \texttt{merge}(-\mathbf{X}_2, \mathbf{X}_1), \\
&\text{output} = \cos(\boldsymbol{\Theta}) \odot \mathbf{X} + \sin(\boldsymbol{\Theta}) \odot \mathbf{X}_{\text{new}},
\end{aligned}
\label{eq:rope}
\end{equation}
where $\boldsymbol{\Theta} = [\boldsymbol{\theta}_1, \ldots, \boldsymbol{\theta}_s] \in \mathbb{R}^{s \times d/2}$ represents the collection of rotation angles for all 
positions in the sequence.
The \texttt{split} and \texttt{merge} operations have several implementations, which differ in how the feature dimensions are paired:
\begin{itemize}
    \item \textbf{Half:} The input feature is divided into two equal halves, and pairs are defined as $(x_1, x_{d/2+1}), (x_2, x_{d/2 + 2}),\ldots, (x_{d/2}, x_d)$.
    \item \textbf{Interleave:} Adjacent dimensions are paired as $(x_0, x_1), (x_2, x_3), \ldots, (x_{d-1},x_d)$, leading to better cache locality and often higher hardware efficiency (\eg, fewer memory-bound vector operations).
    \item \textbf{Others:} Other implementations include quarter~\cite{shi2025kvlatentdimensionallevelkvcache} and interleave-half (\eg, in Mistral~\cite{jiang2023mistral7b} and DeepSeek~\cite{Liu2024}). We include their details in Appendix.
\end{itemize}
These variants preserve the mathematical structure of RoPE while balancing efficiency and representational diversity.

\input{fig/fig2}

\paragraph{2D and 3D Extensions.}
In multimodal and generative models, RoPE has been extended beyond 1D token sequences to encode spatial and temporal structures. This is typically achieved by partitioning the hidden dimension as $d = d_t + d_h + d_w$ and applying RoPE independently along the temporal ($d_t$), height ($d_h$), and width ($d_w$) axes. The 2D case can be derived analogously by omitting the temporal axis. This factorized design enables RoPE to model spatial and temporal dependencies in a separable manner, improving motion consistency and maintaining equivariance to local spatiotemporal shifts. Such 3D RoPE variants are now widely adopted in diffusion-based video generators and vision–language models~\cite{ma2025stepvideot2vtechnicalreportpractice,kong2025hunyuanvideosystematicframeworklarge,wan2025,labs2025flux}, providing unified positional modeling across modalities. However, the repeated division and concatenation of hidden dimensions introduce substantial computational overhead on modern NPUs. To address this, we propose a mathematically equivalent yet computationally efficient alternative that \textbf{replaces these costly data operations with a structured rotation matrix}.

%% file: fig/fig2.tex
\begin{figure*}[!t]
\centering
\includegraphics[width=\linewidth]{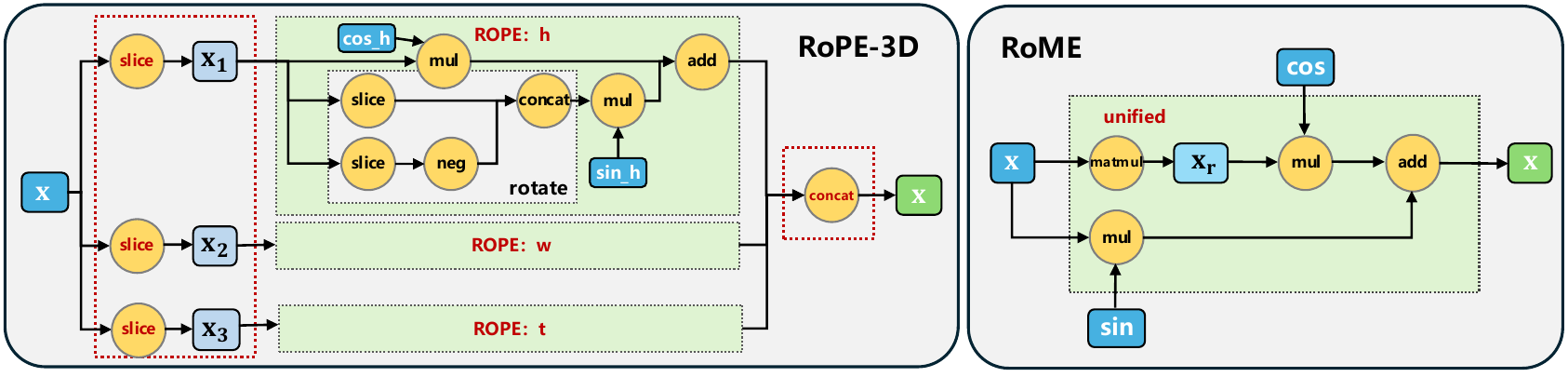}
\vspace{-4mm}
\caption{Illustration of our efficient matrix RoPE implementation (RoME). RoME replaces the complex and memory-intensive operations commonly used in various standard RoPE implementations, like \texttt{rearrange}, \texttt{chunk}, \texttt{cat}, and \texttt{flatten}, with a single structured rotation matrix $\bM$. This unified matrix formulation substantially accelerates both small and fused operators, while also simplifying the overall implementation.}
\label{figure:RoME}
\end{figure*}


%% file: sec/4_method.tex
\section{RoME: An Efficient and Generalizable RoPE Implementation.}
We propose \textbf{RoME}, an efficient and generalizable implementation of Rotary Position Embedding (RoPE) for Transformer training and inference. 
RoME streamlines the computation in four steps: 1. \textbf{Matrix-based transformation.} 
The costly data operations in standard RoPE implementations—such as \texttt{rearrange}, \texttt{chunk}, \texttt{cat}, and \texttt{flatten}—are replaced by a single efficient matrix multiplication.
2. \textbf{Unified multidimensional formulation.} 
For higher-dimensional variants (e.g., RoPE-2D and RoPE-3D), RoME introduces a mathematically equivalent matrix composition method that removes repeated slicing and concatenation, providing a unified and efficient formulation.
3. \textbf{Operator fusion.} 
The sequence of element-wise \texttt{mul}, \texttt{add}, and \texttt{mul} operations is fused into one composite operator (\texttt{mul\_add\_mul}), reducing memory transfers and kernel fragmentation.
4. \textbf{Cube--Vector co-parallel computation.} 
Matrix multiplications (Steps~1--2) are executed on Cube units, while the fused operator (Step~3) runs on Vector units. We execute them in parallel for accelerated training and inference.

\subsection{Matrix-based RoPE transformation}
As shown in Figure \ref{figure:RoME}, we observe that the costly data operations in standard RoPE implementations can be compactly expressed as a linear transformation:
\begin{equation}
\mathbf{x}_{\text{new}} = \mathbf{M} \mathbf{x},
\end{equation}
where $\mathbf{M}$ is a matrix. For the interleave RoPE implementation, $\bM$ is a is a block-diagonal matrix composed of $2\times2$ rotation generators:
\begin{equation}
\mathbf{M} = 
\begin{bmatrix}
0 & 1 \\
-1 & 0
\end{bmatrix}_{\!\text{block diag}},
\end{equation}
For the half RoPE implementation:
\begin{equation}
\mathbf{M} = 
\begin{bmatrix}
\bf{0} & \bI \\
-\bI & 0
\end{bmatrix},
\end{equation}
where the shape of $\bI$ is $d/2 \times d/2$. Please find $\bM$ for other implementation in Appendix.

Substituting into the RoPE formulation, we obtain:
\begin{equation}
\text{output} = \cos \bm{\theta}\bx  + \sin \bm{\theta} \, \mathbf{M} \mathbf{x}.
\label{eq:RoME}
\end{equation}
This reveals that RoPE is equivalent to applying a parameterized orthogonal transformation, which completely eliminates costly slicing operations.
Compared with the standard RoPE formulation (Eq.~\ref{eq:rope}), our matrix-based design 
performs the rotation independently at each position, rather than jointly across all positions $\boldsymbol{\Theta}$. This brings several advantages:
\begin{itemize}
    \item \textbf{Elimination of slicing overhead.} The operations of \texttt{chunk} and \texttt{cat} are replaced by a single matrix multiplication, avoiding memory fragmentation.
    \item \textbf{Better parallelization.} The structure of $\mathbf{M}$ allows batched matrix multiplication, leveraging hardware acceleration efficiently.
    \item \textbf{Mathematical clarity.} The rotation is expressed explicitly as a linear operator, making theoretical extensions (e.g., higher-dimensional RoPE, video spatio-temporal RoPE) easier to formulate.
\end{itemize}

\subsection{Unified multidimensional formulation}
For higher-dimensional inputs such as videos or spatial-temporal features, conventional RoPE applies rotary encoding independently along each divided dimensions (\eg, height, width, and temporal), resulting in multiple rounds of slicing, reshaping, and rotation. 
This processing introduces redundant data movement and kernel launches, limiting computational efficiency on large tensors.

In contrast, \textbf{RoME} reformulates multidimensional rotary embeddings into a unified matrix representation. 
Instead of performing separate rotations for each axis, RoME constructs a block-diagonal matrix that directly combines the corresponding rotation matrices of each dimension:
\begin{equation}
\mathbf{M}_{\text{3D}} = \text{diag}\!\left(\mathbf{M}_t, \mathbf{M}_h, \mathbf{M}_w\right),
\end{equation}
where $\mathbf{M}_t, \mathbf{M}_h, \mathbf{M}_w$ are the rotation matrices in Eq \ref{eq:RoME} on the divided dimensions, encoding temporal, height, and width positional information, respectively. An example is shown in Figure~\ref{figure:3dRope}.
This formulation enables all rotations to be applied simultaneously through a single matrix multiplication, without costly tensor slicing and concatenation operations.


Furthermore, this approach naturally generalizes to higher-dimensional cases, where RoME constructs
\begin{equation}
\mathbf{M}_{\text{nD}} = \text{diag}\!\left(\mathbf{M}_1, \mathbf{M}_2, \dots, \mathbf{M}_n\right),
\end{equation}
for $n$-dimensional positional encodings.
By treating each dimension’s positional rotation as an independent block on the diagonal, RoME provides a mathematically consistent and hardware-efficient formulation that maintains the expressive power of multidimensional rotary embeddings while minimizing memory access overhead.

\input{fig/fig3}

\subsection{Operator fusion for performance optimization}
According to Eq.~\ref{eq:RoME}, the mathematical formulation of rotary position embedding involves two multiplications and one addition.
By applying operator fusion, these three operations can be combined into a single fused operator, \texttt{mul\_add\_mul}, which significantly improves computational efficiency.

\subsection{Cube Vector co-parallel computation}
After applying our proposed scheme, RoPE involves only two operators: \texttt{matmul} and \texttt{mul\_add\_mul}.
Among them, \texttt{matmul} is executed on the Cube unit, while \texttt{mul\_add\_mul} runs on the Vector unit.
As shown in Figure~\ref{figure:parallel}, by properly scheduling the computation pipeline, these two operations can be processed in parallel, effectively improving hardware utilization. 

\input{fig/fig4}

%% file: fig/fig3.tex
\begin{figure}[!t]
\centering
\includegraphics[width=\linewidth]{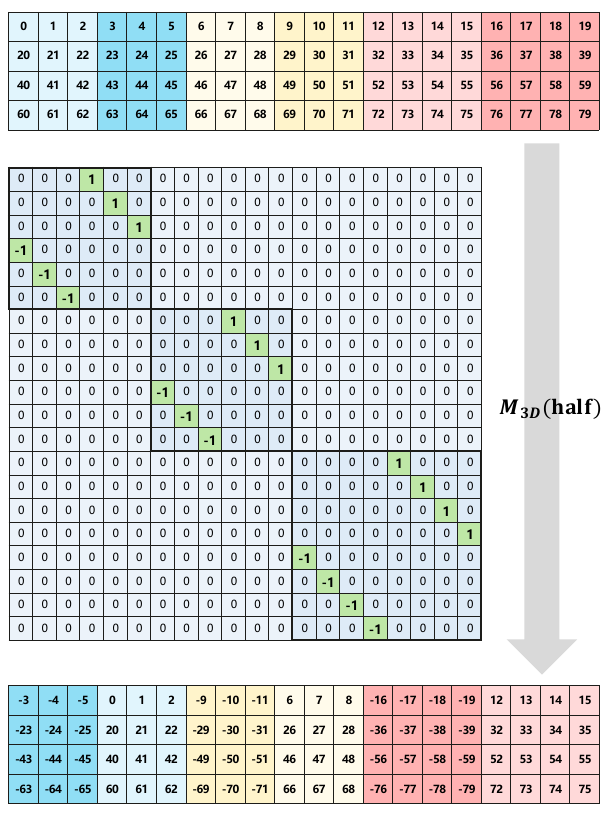}
\caption{An example of transforming input $\mathbf{x}$ by $\mathbf{M}_{\text{3D}}$. Sequence length and hidden dimension is 4 and 20, respectively.}
\label{figure:3dRope}
\vspace{-4mm}
\end{figure}

%% file: fig/fig4.tex
\begin{figure}[!t]
\centering
\includegraphics[width=\linewidth]{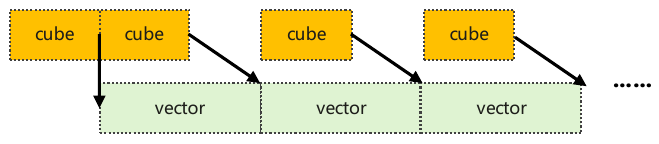}
\caption{Illustration of our Cube Vector co-parallel computation.}
\label{figure:parallel}
\end{figure}

%% file: sec/5_exp.tex
\section{Experiments}

\subsection{Settings}

\paragraph{Setup.} 
We evaluate RoME under both operator-level and model-level settings. The following configurations are used unless otherwise specified:

\begin{itemize}[leftmargin=1em]
\setlength{\itemsep}{2pt}
\setlength{\parsep}{2pt}
\setlength{\parskip}{2pt}

\item[*] \textbf{Environment.} Experiments are conducted on 8$\times$ Ascend 910B3 NPUs (64\,GB each) using the CANN and PyTorch environments. Single-operator and lightweight model tests can also be run on a single card if memory permits.

\item[*] \textbf{Dataset.} Since our focus is acceleration rather than accuracy, we randomly sample data from real-world datasets. Typical preprocessing steps are applied to align inputs with hardware-friendly dimensions (\eg, resizing video inputs from 270P to 272P).

\item[*] \textbf{Model.} We benchmark representative models such as Wan 2.2~\cite{wan2025} under MindSpeed-MM, MindSpeed-LLM, and VLLM frameworks. For single-operator evaluation, the testing code is available in the open-source repository \texttt{gitcode/cann/ops-transformer/}. Integration with these frameworks requires \texttt{torch.compile} or our custom \texttt{pybind} interface.

\item[*] \textbf{Shape configuration.} Unless stated otherwise, we set the batch size to $B{=}1$ per card and head dimension $D{=}128$. For single-operator experiments, we use the representative configuration $[B, N, S, D] = [1, 24, 28800, 128]$, where $B$ denotes batch size, $N$ the number of attention heads, $S$ the sequence length, and $D$ the head dimension.
\end{itemize}



\paragraph{Baselines.} 
We compare RoME against standard RoPE implementations from widely used open-source frameworks~\cite{mindSpeed2025,vllmAscend2025}. Most implementations follow Equation~\ref{eq:rope}, combining PyTorch low-level APIs for splitting, rotating, and merging features. In cases where fused kernels are available, we use these optimized versions as stronger baselines to ensure a fair comparison. For model-level benchmarks, we use TTFT(Time to First Token~\cite{grattafiori2024llama3herdmodels}) as the baseline for LLMs, and the one-step inference latency as the baseline for video generation models.

\paragraph{Metrics.} 
We report both \emph{speedup times} $t_0/t$ and \emph{speedup percentage} $(t_0-t)/t_0$, where $t_0$ denotes the baseline runtime and $t$ denotes the optimized runtime using RoME.

\input{tables/rope_performance}

\input{tables/infer_performance}

\input{tables/train_performance}


\subsection{Main results}


Table~\ref{tab:rope_performance} summarizes the single-operator results, showing that RoME consistently outperforms standard RoPE implementations across most configurations. We then integrate RoME into several widely used models and evaluate it on multiple tasks. As shown in Table~\ref{tab:infer_performance}, RoME delivers a $3\%\text{–}10\%$ inference speedup in most cases. An exception is the \emph{Hunyuan Video} model, where the observed speedup is below $1\%$. Profiling reveals that this model already employs heavily optimized, case-specific fused kernels with a 1D RoPE implementation, leaving little headroom for further acceleration. In such scenarios, the key advantage of RoME lies in its \textbf{generality}: it requires no deep, model-specific kernel engineering, while still achieving competitive performance compared to highly specialized fused operators.

As shown in Table~\ref{tab:train_performance}, training also benefits from RoME, since the backward pass faces the same bottlenecks as the forward pass. For example, applying RoME to the \texttt{wan2.2/1.3B} model yields clear end-to-end training improvements, further supporting the effectiveness of our approach.

\input{tables/complexity}

\input{tables/ablation}

\subsection{Ablation}
\label{sec:5.3}

\paragraph{Theoretical and empirical complexity analysis.} 
Let the hidden dimension be $d$ and the sequence length be $s$. In the conventional RoPE implementation, the rotation is realized through a sequence of vector-level operations such as elementwise addition, interleaving, and concatenation. Although these operations have an arithmetic complexity of $\mathcal{O}(s d)$, they require multiple kernel launches and incur substantial memory I/O, which often becomes the dominant cost in practice.
In contrast, our matrix-based formulation replaces these vector operations with a single matrix multiplication:
\begin{equation}
\mathbf{Y} = \mathbf{R} \mathbf{X}, \qquad 
\mathbf{R} = \cos \bm{\theta}\, \mathbf{I} \;+\; \sin \bm{\theta}\, \mathbf{M},
\end{equation}
where $\mathbf{R}$ is a structured block-diagonal rotation matrix derived from the RoPE angles. This approach maintains the same theoretical arithmetic complexity of $\mathcal{O}(s d)$ but significantly reduces memory movement and allows for effective kernel fusion. As a result, the matrix-based implementation achieves empirically faster execution on both Ascend(NPU) and GPU architectures.


\paragraph{Breakdown of Various Implementations.}
Table~\ref{tab:complexity} presents a theoretical and empirical comparison of three RoPE implementations:  
(1) the conventional \texttt{chunk+cat} design,  
(2) our matrix-based formulation using a block-diagonal rotation matrix $\mathbf{M}$, and  
(3) the fully explicit rotation approach that constructs a dense $d \times d$ matrix $\mathbf{R}$.
The explicit rotation method directly instantiates $\mathbf{R} = \cos\bm{\theta}\,\mathbf{I} + \sin\bm{\theta}\,\mathbf{M}$ and applies a dense matrix multiplication, which substantially increases both arithmetic cost and memory traffic. In contrast, our block-diagonal formulation achieves the same linear rotation behavior while maintaining $\mathcal{O}(d)$ complexity and avoiding the excessive kernel launches associated with \texttt{chunk+cat}. As the table shows, this structure significantly reduces overhead in practice, leading to superior empirical efficiency.

\paragraph{Module-level Ablations for RoME.}
To better understand the contribution of each component in RoME, we decompose the method into three parts:  
(1) the mathematically equivalent matrix-based transformation,  
(2) the natural ability to fuse 2D/3D positional encoding into a single 1D operation, and  
(3) fully or partially fused kernel optimizations.
We conduct profiling on 3D interleaved RoPE using the original shape $[1,24,28800,128]$ and its decomposed components: 
$h{=}[1,24,28800,44]$, $w{=}[1,24,28800,44]$, and $t{=}[1,24,28800,40]$.
As shown in Table~\ref{tab:ablation}, the matrix-based replacement provides consistent improvements across all settings. The fused operator and 3D→1D conversion yield additional speedups for certain combinations. However, without converting 3D RoPE to its 1D equivalent, fused operators may occasionally degrade performance, since some 3D RoPE implementations are difficult to map exactly to a 1D formulation~\cite{ma2025stepvideot2vtechnicalreportpractice}.  
Overall, RoME achieves the best performance when all components are applied together, demonstrating strong efficiency gains.

%% file: tables/rope_performance.tex
\begin{table*}[ht!]
    \centering
    \caption{Time taken for a single operator (applying position embedding) and speedup times for RoME compared with different RoPE implementations. RoPE refers to it's origin application and 2D/3D sequels~\cite{Su2021RoFormerET,heo2024rotarypositionembeddingvision,ma2025stepvideot2vtechnicalreportpractice}. RoPE+fuse refers to \emph{model-specific} optimization implemented in~\citep{mindSpeed2025,vllmAscend2025}, where 2D and 3D operators are fused for efficient computation.}
    \label{tab:rope_performance}
    \resizebox{0.9\textwidth}{!}{
    \begin{tabular}{|c|c|c|c|c|c|c|}
    \hline
    Mode & Dimension & RoPE & RoPE+fuse &\textbf{RoME} & \textbf{vs. RoPE} & \textbf{vs. RoPE+fuse} \\
    \hline
    \multirow{2}{*}{interleave} & 2D (dim=64+64) & 5.2ms & \multirow{2}{*}{\shortstack{2.0ms}} & \multirow{4}{*}{\textbf{1.6ms}} & 3.3x & \multirow{2}{*}{1.3x} \\
    \cline{2-3} \cline{6-6}
     & 3D (dim=44+44+40) & 5.7ms & & & 3.6x & \\
    \cline{2-4} \cline{6-7}
    \multirow{2}{*}{half} & 2D (dim=64+64) & 5.0ms & 2.2ms & \multirow{2}{*}{} & 2.1x & 1.4x \\
    \cline{2-4} \cline{6-7}
     & 3D (dim=44+44+40) & 5.8ms & 9.9ms & & 3.6x & 6.2x \\
    \hline
    \end{tabular}}
    \end{table*}


%% file: tables/infer_performance.tex
\begin{table*}[ht!]
    \centering
    \caption{Acceleration in inference by applying RoME vs. RoPE.}
    \label{tab:infer_performance}
    \scalebox{0.9}{
    \begin{tabular}{|c|c|c|c|c|c|c|}
    \hline
    {Task} & Model & Parameters & QKShape[BNSD] & Baseline(ms)  & \textbf{RoME(ms)} & \textbf{Speedup}  \\
    \hline
    LLM           & Llama3.1       & 8B                  & $[1, 24, 8192,128]$    & $50.8$                  & $48.1$           & \cellcolor{mygray}\textbf{$4.7\%$}\\
    \hline
    VLM             & internvl3       & 8B          & $[1, 28, 1846, 128]$    & $174$                       & $166$            & \cellcolor{mygray}\textbf{$4.5\%$}\\
    \hline
    Edit            & FLUX.1-Kontext      & 12B          & $[1, 24, 8704, 128]$    & $1059$                  & $978$            & \cellcolor{mygray}\textbf{$8.3\%$}\\
    \hline
    T2I             & Qwen-Image       & 20B          & $[1, 24, 6032, 128]$    & $1457$                  & $1401$              & \cellcolor{mygray}\textbf{$4\%$}\\
    \hline
    \multirow{6}{*}{T2V} & \multirow{2}{*}{HunyuanVideo}    &\multirow{2}{*}{13B}     & $[1, 24, 42840, 128]/540P$    & $14410$  & $14400$  & \cellcolor{mygray}\textbf{$0.1\%$}\\
                                                                                \cline{4-7}
                                                                                &&      & $[1, 24, 8192,128]/720P$    & $40120$  & $40100$ & \cellcolor{mygray}\textbf{$0.1\%$}\\

                        \cline{2-7} 
                        & \multirow{4}{*}{Wan2.2}    &\multirow{2}{*}{1.3B}     & $[1, 12, 6630, 128]/272P$    & $209$    & $199$    & \cellcolor{mygray}\textbf{$3.9\%$}\\
                                                                                \cline{4-7}
                                                                                &&      & $[1, 12, 17550, 128]/480P$    & $790$     & $759$   & \cellcolor{mygray}\textbf{$3.4\%$}\\
                                                    \cline{3-7} 
                        &                           &\multirow{2}{*}{14B}        & $[1, 40, 6630, 128]/272P$    & $1170$    & $1120$    & \cellcolor{mygray}\textbf{$4.3\%$}\\
                                                                                \cline{4-7}
                                                                                &&      & $[1, 40, 17550, 128]/480P$    & $4500$     & $4350$   & \cellcolor{mygray}\textbf{$3.3\%$}\\
    \hline
    \end{tabular}}
\end{table*}

%% file: tables/train_performance.tex
\begin{table}[h!]
    \centering
    \caption{Time taken for each training step using RoPE vs RoME. Model is \texttt{wan2.2/1.3B}.}
    \label{tab:train_performance}
    \scalebox{0.9}{
    \begin{tabular}{lccc}
    \hline
    {Inputs} & RoPE (s) & \textbf{RoME (s)} & \cellcolor{mygray}\textbf{Speedup}  \\
    \hline
    $272P$           & $1.27$      & \textbf{$1.22$}  & \cellcolor{mygray}$4\%$\\
    \hline
    $480P$           & $3.37$      & \textbf{$3.30$}  & \cellcolor{mygray}$2.3\%$\\
    \hline
    \end{tabular}}
\end{table}

%% file: tables/complexity.tex
\begin{table*}[t]
    \centering
    \caption{Comparison of RoPE implementations. Our matrix-based approach achieves the best trade-off between efficiency and simplicity, reducing vector-operator cost that accounts for over 50\% of the forward-time MFU bottleneck.}
    \scalebox{0.8}{
    \begin{tabular}{lccc}
    \hline\hline
      & Chunk+Cat       & \cellcolor{mygray}\textbf{Matrix ($\mathbf{M}$)} & Direct Rotation ($\mathbf{R}$) \\
    \hline
    Algorithmic complexity             & $\mathcal{O}(d)$ & \cellcolor{mygray}$\mathcal{O}(d)$               & $\mathcal{O}(d^2)$ \\
    Memory access pattern              & Non-contiguous   & \cellcolor{mygray}Contiguous                     & Dense matmul \\
    Kernel launches                    & 3--4 ops         & \cellcolor{mygray}\textbf{1 fused op}            & 1 heavy op \\
    Parallel efficiency                & Low              & \cellcolor{mygray}\textbf{High}                  & Moderate \\
    Implementation simplicity          & Moderate         & \cellcolor{mygray}\textbf{High}                  & Low \\
    Fusibility into attention          & No               & \cellcolor{mygray}\textbf{Yes}                   & Difficult \\
    Empirical speed (Ascend 910B3)     & 1.00$\times$     & \cellcolor{mygray}\textbf{$\geq$3.7$\times$}     & 0.01$\times$ \\
    \hline\hline
    \end{tabular}}
    \label{tab:complexity}
\end{table*}


%% file: tables/ablation.tex
\begin{table*}[h!]
    \centering
    \caption{Ablation study for each modules in RoME. Using none of them corresponds to standard RoPE.}
    \scalebox{0.8}{
    \begin{tabular}{p{3cm} p{1cm}<{\centering}p{1cm}<{\centering}p{1cm}<{\centering}p{1cm}<{\centering}p{1cm}<{\centering}p{1cm}<{\centering}p{1cm}<{\centering}p{3cm}<{\centering}}
    \hline\hline
    \multicolumn{1}{l}{\textbf{Method}} & \textbf{RoPE} &  &  & & & & & \textbf{RoME (Ours)} \\
    \hline
    Matrix transfer                    & -        & \checkmark  & -           & -             & \checkmark  & \checkmark  & -           & \checkmark     \\
    Input fusion                       & -        & -           & \checkmark  & -             & \checkmark  & -           & \checkmark  & \checkmark     \\
    Operator fusion                    & -        & -           & -           & \checkmark    & -           & \checkmark  & \checkmark  & \checkmark     \\
    \hline
    Infer Cost(ms)                     & 6.0      & 5.7         & 4.7         & 7.7           & 2.8         & 3.7         & 2.1         & \cellcolor{mygray} \textbf{1.6}  \\
    vs. RoPE                       & -        & 1.1x        & 1.3x        & \textbf{0.8x} & 2.1x        & 1.6x        & 2.8x        & \cellcolor{mygray} \textbf{3.7x} \\
    \hline \hline
    \end{tabular}}
    \label{tab:ablation}
    \vspace{-0.3cm}
    \end{table*}

%% file: sec/6_conclusion.tex

\section{Conclusion}
We presented RoME, a mathematically equivalent yet computationally efficient reformulation of Rotary Position Embedding. By replacing RoPE’s vector-level \texttt{split} and \texttt{merge} operations with a unified matrix transformation, RoME significantly reduces memory movement, improves kernel fusion, and simplifies implementation across a wide range of models. Our analysis shows that many of the inefficiencies in existing RoPE implementations arise not from the rotation itself but from the surrounding dataflow, particularly in multi-dimensional (2D/3D) settings. RoME addresses these bottlenecks in a principled manner, enabling efficient execution on both Ascend(NPUs) and GPUs architectures. Our matrix-based formulation also opens avenues for further generalizing RoPE to richer multi-dimensional modalities, including long-context video generation, 3D scene understanding, and spatiotemporal foundation models.


%% file: sec/X_suppl.tex
\clearpage
\setcounter{page}{1}
\maketitlesupplementary

\renewcommand{\thesection}{A.\arabic{section}}
\renewcommand*{\thesubsection}{A.\arabic{section}.\arabic{subsection}}
\renewcommand{\thetable}{A\arabic{table}}
\renewcommand{\thefigure}{A\arabic{figure}}
\renewcommand{\theequation}{A\arabic{equation}}


\section{Different Type of RoME Implement }
\textbf{Algorithms}: $\forall$RoPE, $\exists$ $(d,d)-size$ Matrix $M$ to make it mathematical equivalent to RoME(Eq.~7).
Thus, the key for RoME is to find the $(d,d)-size$ Matrix M. 
Unless otherwise specified, the RoME satisfy Eq.~7.

\subsection{1D-RoME}
\paragraph{Half.} As mention in main body, half-mode RoPE can be written as Eq.~\ref{eq:rope_half}
\begin{equation}
    \begin{aligned}
    &x_1, x_2 = torch.chunk(x, 2, dim=-1) \\
    &x_{new} = torch.cat((-x_2, x_1), dim=-1) \\
    &rope_{out} = x * cos + x_{new} * sin
    \end{aligned}
    \label{eq:rope_half}
    \end{equation}
For this mode, we can find a matrix M:
\begin{equation}
\mathbf{M} = 
\begin{bmatrix}
\bf{0} & \bI \\
-\bI & 0
\end{bmatrix},
\end{equation}
where the shape of $\bI$ is $d/2 \times d/2$.
\begin{figure}[!t]
    \centering
    \includegraphics[width=\linewidth]{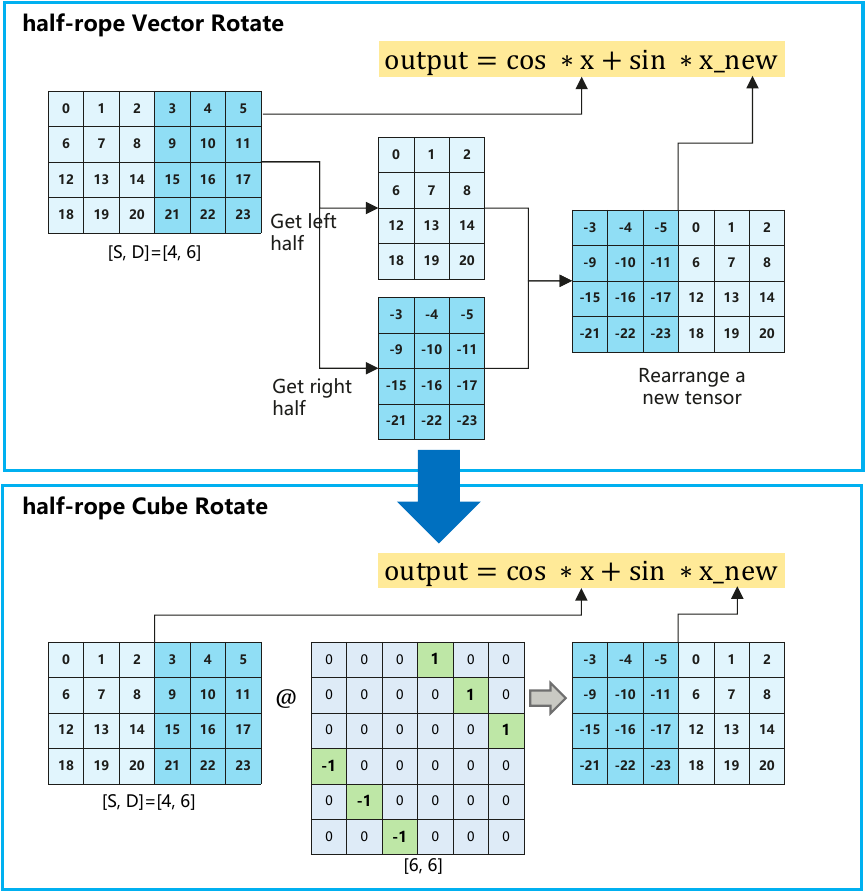}
    \caption{An example of RoPE-half.}
    \label{figure:rope_half}
\end{figure}

\paragraph{Interleave.} Similar as half mode:
\begin{equation}
    \begin{aligned}
    &{\color{blue}x = rearrange(x, ... (d j) -> ... d j, j=2)} \\
    &x_1, x_2 = torch.chunk(x, 2, dim=-1) \\
    &x_{new} = torch.cat((-x_2, x_1), dim=-1).flatten(-2) \\
    &rope_{out} = x * cos + x_{new} * sin
    \end{aligned}
    \label{eq:rope_interleave}
\end{equation}
For this mode, $\bM$ is a is a block-diagonal matrix composed of $2\times2$ rotation generators:
\begin{equation}
    \mathbf{M} = 
    \begin{bmatrix}
    0 & 1 \\
    -1 & 0
    \end{bmatrix}_{\!\text{block diag}},
\end{equation}

\begin{figure}[!t]
    \centering
    \includegraphics[width=\linewidth]{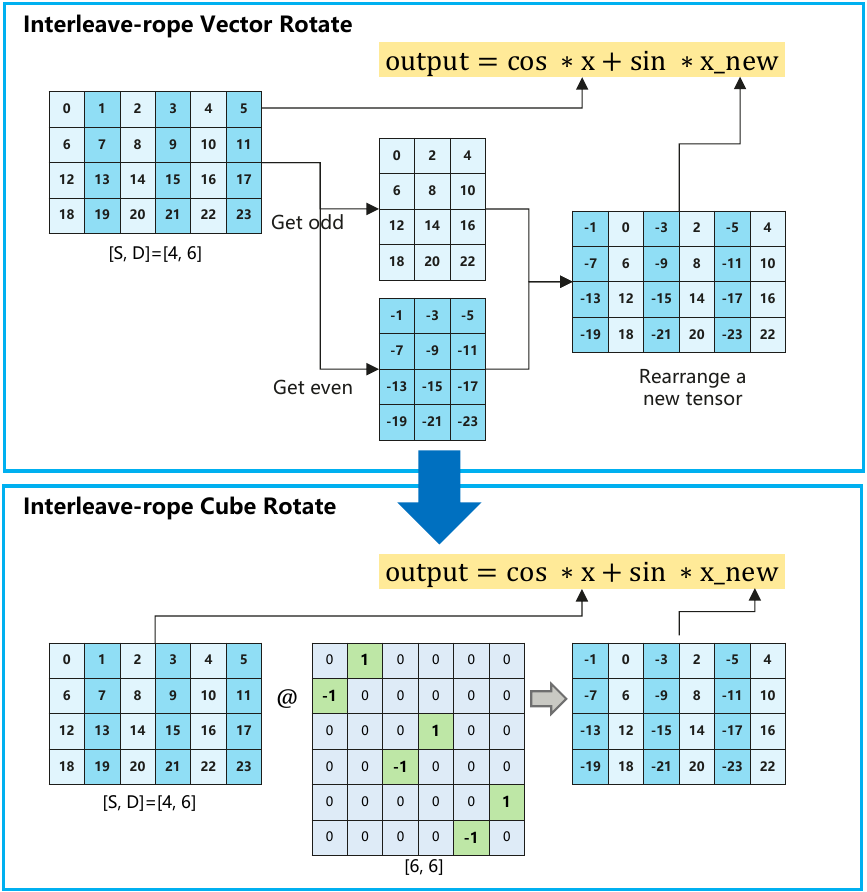}
    \caption{An example of RoPE-interleave.}
    \label{figure:rope_interleave}
\end{figure}

\paragraph{Interleave-Half.}
This mode combines the ideas of \emph{interleave} and \emph{half}.  
First, all even-indexed dimensions are moved to the first half of the feature vector, and all odd-indexed dimensions to the second half.  
The resulting vector is then processed in the same manner as in the \texttt{half} mode.  
Formally:

\begin{equation}
\label{eq:rope_interleave_half}
\begin{aligned}
    &x = \text{rearrange}(x,\; \ldots (d\, j) \rightarrow \ldots\, d\, j,\; j{=}2) \\
    &x_1, x_2 = \text{torch.chunk}(x,\; 2,\; \text{dim}{=}{-}1) \\
    &\phantom{x_1, x_2 =}\;.\text{squeeze}(\text{dim}{=}{-}1) \\
    &\color{blue}{x = \text{torch.cat}((x_1, x_2),\; \text{dim}{=}{-}1)} \\
    &x_{\text{new}} = \text{torch.cat}((-x_2,\, x_1),\; \text{dim}{=}{-}1) \\
    &\text{rope}_{\text{out}} = x \cdot \cos + x_{\text{new}} \cdot \sin
\end{aligned}
\end{equation}

This mode is a special case and cannot be directly expressed using Eq.~7.  
However, we can derive an analogous formulation using the matrix-based viewpoint, which we denote as the \textbf{RoME-extension}:

\begin{equation}
\label{eq:RoME_EXT}
    \text{output}
    = \cos \bm{\theta} \, \mathbf{M_1} \mathbf{x}
      + \sin \bm{\theta} \, \mathbf{M_2} \mathbf{x}.
\end{equation}

Here, $\mathbf{M_1}$ is a $(d \times d)$ permutation matrix, and $\mathbf{M_2}$ is a modified permutation matrix in which half of the entries $1$ are replaced with $-1$. Their definitions are:

\begin{equation}
\mathbf{M_1} =
\left\{
\begin{array}{ll}
    m_{ij} = 1, & j < \frac{d}{2},\;\; i = 2j, \\
    m_{ij} = 1, & j \ge \frac{d}{2},\;\; i = 2(j-\frac{d}{2}) + 1, \\
    m_{ij} = 0, & \text{otherwise},
\end{array}
\right.
\end{equation}

\begin{equation}
\mathbf{M_2} =
\left\{
\begin{array}{ll}
    m_{ij} = -1, & j < \frac{d}{2},\;\; i = 2j + 1, \\
    m_{ij} = 1,  & j \ge \frac{d}{2},\;\; i = 2(j-\frac{d}{2}), \\
    m_{ij} = 0,  & \text{otherwise},
\end{array}
\right.
\end{equation}
where $m_{ij}$ denotes the element at row $0 \le i < d$ and column $0 \le j < d$.

\bigskip

If $\frac{d}{2}$ is a chip-friendly size, the RoME-extension can be expressed more efficiently as:

\begin{equation}
\label{eq:RoME_EXT_V2}
\begin{aligned}
    &\mathbf{x_1},\; \mathbf{x_2} = \mathbf{M_{11}} \mathbf{x},\; \mathbf{M_{12}} \mathbf{x} \\[2pt]
    &\text{output}_1 = \cos_1 \bm{\theta}\; \mathbf{x_1} - \sin_1 \bm{\theta}\; \mathbf{x_2} \\
    &\text{output}_2 = \cos_2 \bm{\theta}\; \mathbf{x_2} + \sin_2 \bm{\theta}\; \mathbf{x_1} \\
    &\text{output} = \text{cat}(\text{output}_1,\; \text{output}_2)
\end{aligned}
\end{equation}
where $\mathbf{M_{11}}$ and $\mathbf{M_{12}}$ correspond to the left and right halves of the columns of $\mathbf{M_1}$, respectively.  
Similarly, $\cos_1,\cos_2,\sin_1,\sin_2$ are the corresponding halves of $\cos$ and $\sin$.  
The final \texttt{cat} operation incurs no overhead because $\frac{d}{2}$ is a hardware-friendly dimension, making $\text{output}_1$ and $\text{output}_2$ contiguous in memory.
\begin{figure*}[!t]
    \centering
    \includegraphics[width=\linewidth]{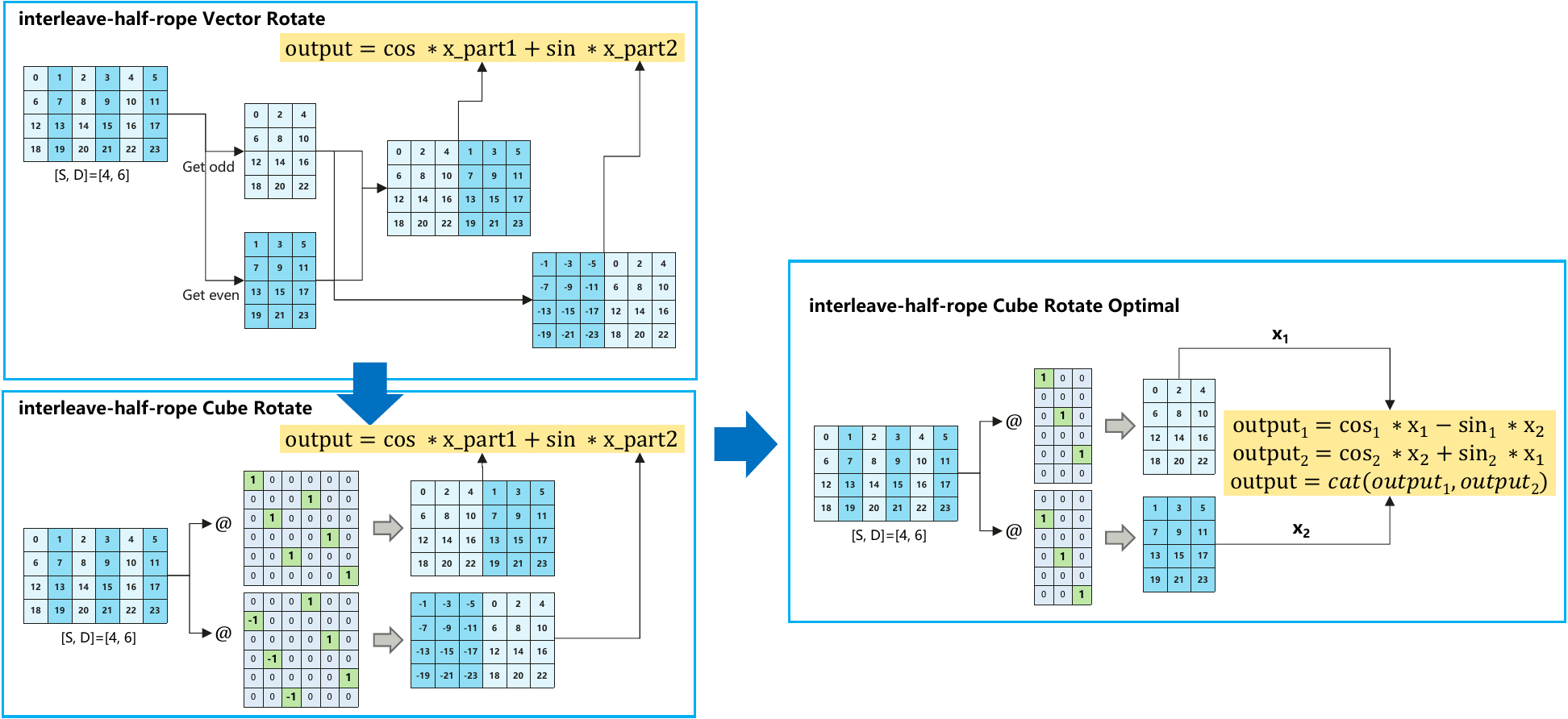}
    \caption{An example of RoPE-interleave-half.}
    \label{figure:rope_interleave_half}
\end{figure*}

\paragraph{Quarter.} Similar as half mode:
\begin{equation}
    \begin{aligned}
        &x_1, x_2, x_3, x_4 = torch.chunk(x, 4, dim=-1) \\
        &x_{new} = torch.cat((-x_2, x_1, -x_4, x_3), dim=-1) \\
        &rope_{out} = x * cos + x_{new} * sin
    \end{aligned}
    \label{eq:rope_quarter}
\end{equation}
For this mode, we can find a matrix M:
\begin{equation}
\mathbf{M} = 
\begin{bmatrix}
\bf{0}   & \bI    &\bf{0} &\bf{0} \\
-\bI     &\bf{0}  &\bf{0} &\bf{0}  \\
\bf{0}   &\bf{0}  &\bf{0} &\bI   \\
\bf{0}   &\bf{0}  &-\bI   &\bf{0}  \\
\end{bmatrix},
\end{equation}
where the shape of $\bI$ is $d/4 \times d/4$.
\begin{figure}[!t]
    \centering
    \includegraphics[width=\linewidth]{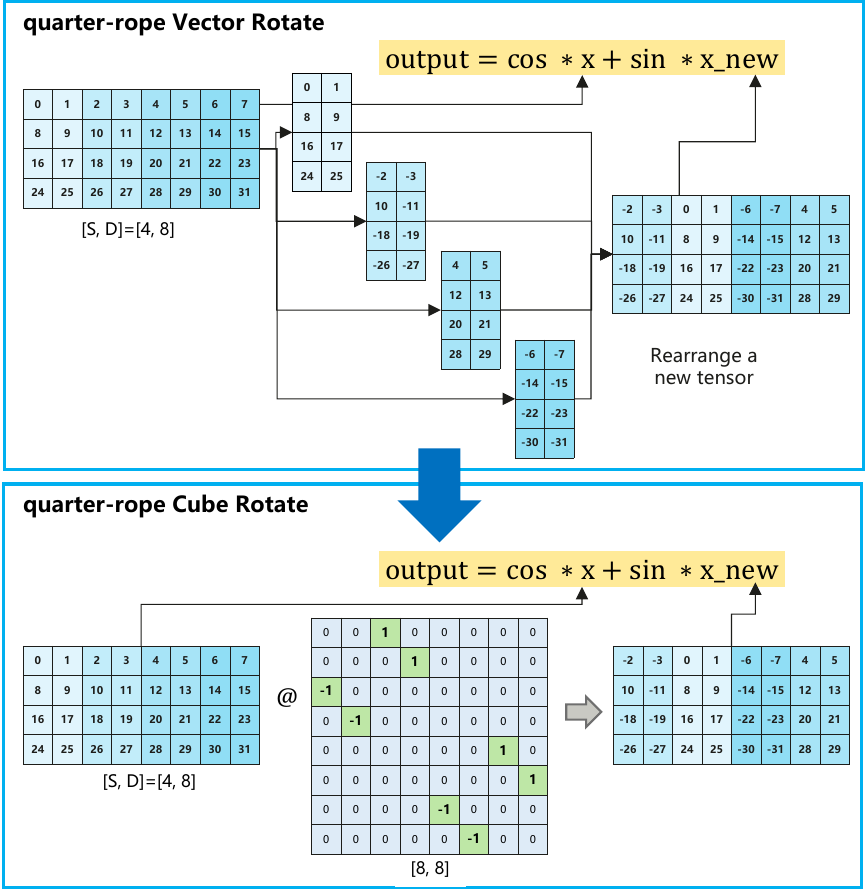}
    \caption{An example of RoPE-quarter.}
    \label{figure:rope_quarter}
\end{figure}

\paragraph{Extensions.} Besides with the modes mentioned in pre-paragraph, there are new type of RoPE designing, e.g., GLM-4.6. 
If want better performance, tranditional way is to write new fused operators for each new design.
However, for RoME, we can get a proper performance with a new $M$ designing, no need engineering problem.  

\subsection{3D-RoME}

In this section, we introduce \textbf{3D-RoME}.  
(2D-RoME can be viewed as a special case of 3D-RoME where the temporal dimension $t$ has size zero.)

\begin{figure*}[!t]
    \centering
    \includegraphics[width=\linewidth]{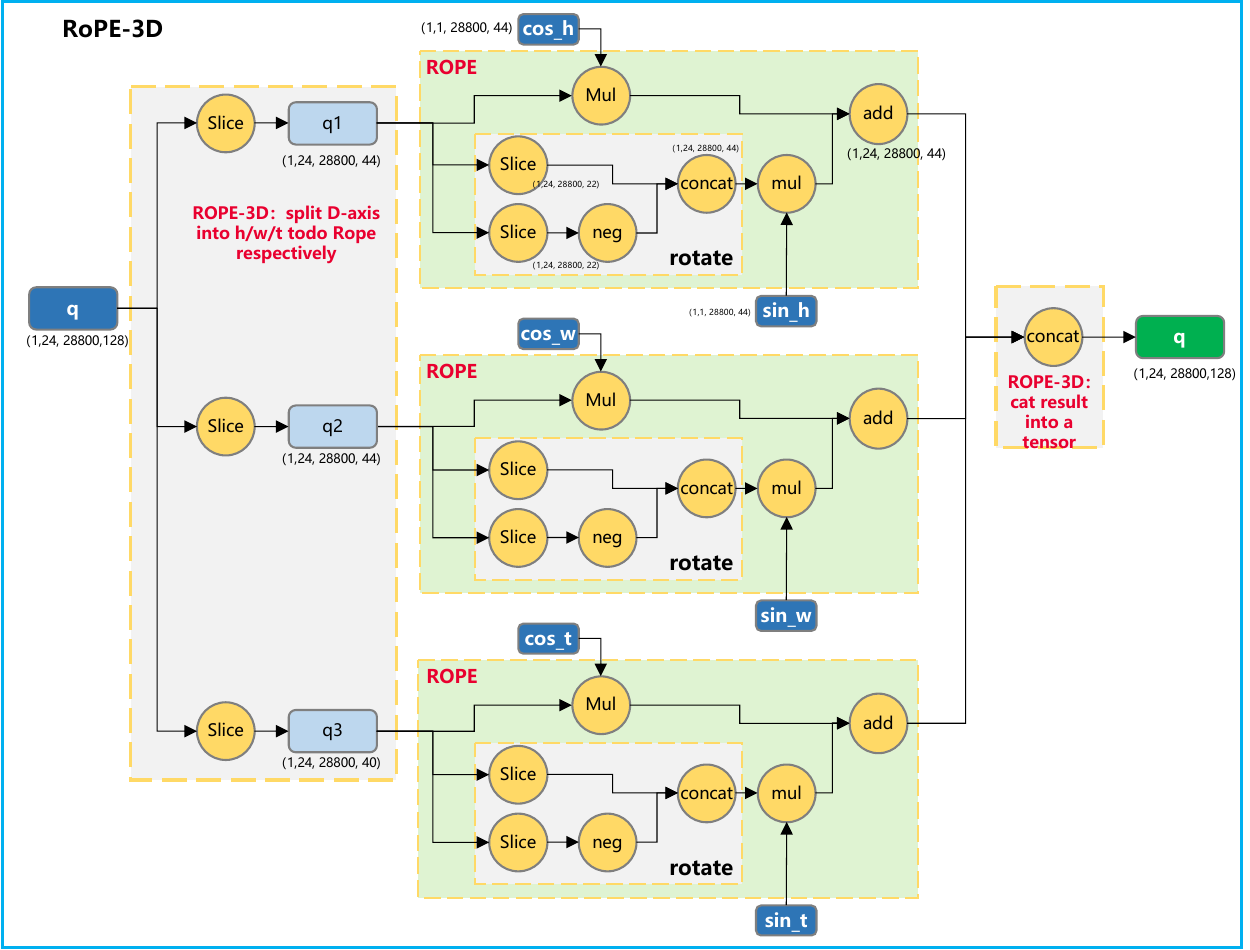}
    \caption{An example of RoPE-3D algorithm.}
    \label{figure:rope_3d}
\end{figure*}

A naive implementation of 3D-RoPE applies standard 1D-RoPE independently along the height, width, and temporal dimensions ($h/w/t$).  
In theory, this appears to have the same arithmetic cost as 1D-RoPE, since the hidden dimension decomposes as $d = d_h + d_w + d_t$.  
However, in practice, 3D-RoPE introduces \emph{additional slicing and concatenation operations}.  
Moreover, although $d$ is typically chosen to be a chip-friendly dimension, the sub-dimensions $d_h$, $d_w$, and $d_t$ are often \emph{not} hardware-friendly, resulting in inefficient kernels and degraded performance.
Therefore, it is desirable to design an algorithm that maintains full compatibility with the chip-friendly dimension $d$.  
Our proposed \textbf{3D-RoME} satisfies this requirement exactly.  
As shown in Fig.~\ref{figure:rope_3d}, the original 3D-RoPE implementation calls the RoPE operator three separate times and contains many hardware-unfriendly operations.  
Using the RoME formulation of Eq.~7, the only remaining challenge is constructing the appropriate structured matrix~$\mathbf{M}$.
Consider the two common RoPE modes—\emph{half} and \emph{interleave}.  
Since the interleave mode simply swaps neighboring even and odd dimensions, it is noteworthy that the 3D version of $\mathbf{M}$ is \emph{identical} to the 1D counterpart.  
For the half mode, the 3D-$\mathbf{M}$ can be written as follows:
\begin{equation}
    \mathbf{M} = 
    \begin{bmatrix}
    \bf{0} & \bI_h  & \bf{0} & \bf{0} & \bf{0} & \bf{0} \\
    -\bI_h & \bf{0} & \bf{0} & \bf{0} & \bf{0} & \bf{0} \\
    \bf{0} & \bf{0}  & \bf{0} & \bI_w & \bf{0} & \bf{0} \\
    \bf{0} & \bf{0} & -\bI_w & \bf{0} & \bf{0} & \bf{0} \\
    \bf{0} & \bf{0}  & \bf{0} & \bf{0} & \bf{0} & \bI_t \\
    \bf{0} & \bf{0} & \bf{0} & \bf{0} & -\bI_t & \bf{0}
    \end{bmatrix},
    \end{equation}
where the shape of $\bI_h$ is $\frac{d_h}{2} \times \frac{d_h}{2}$, $\bI_w$ is $\frac{d_w}{2} \times \frac{d_w}{2}$, $\bI_t$ is $\frac{d_t}{2} \times \frac{d_t}{2}$.

\begin{figure*}[!t]
    \centering
    \includegraphics[width=\linewidth]{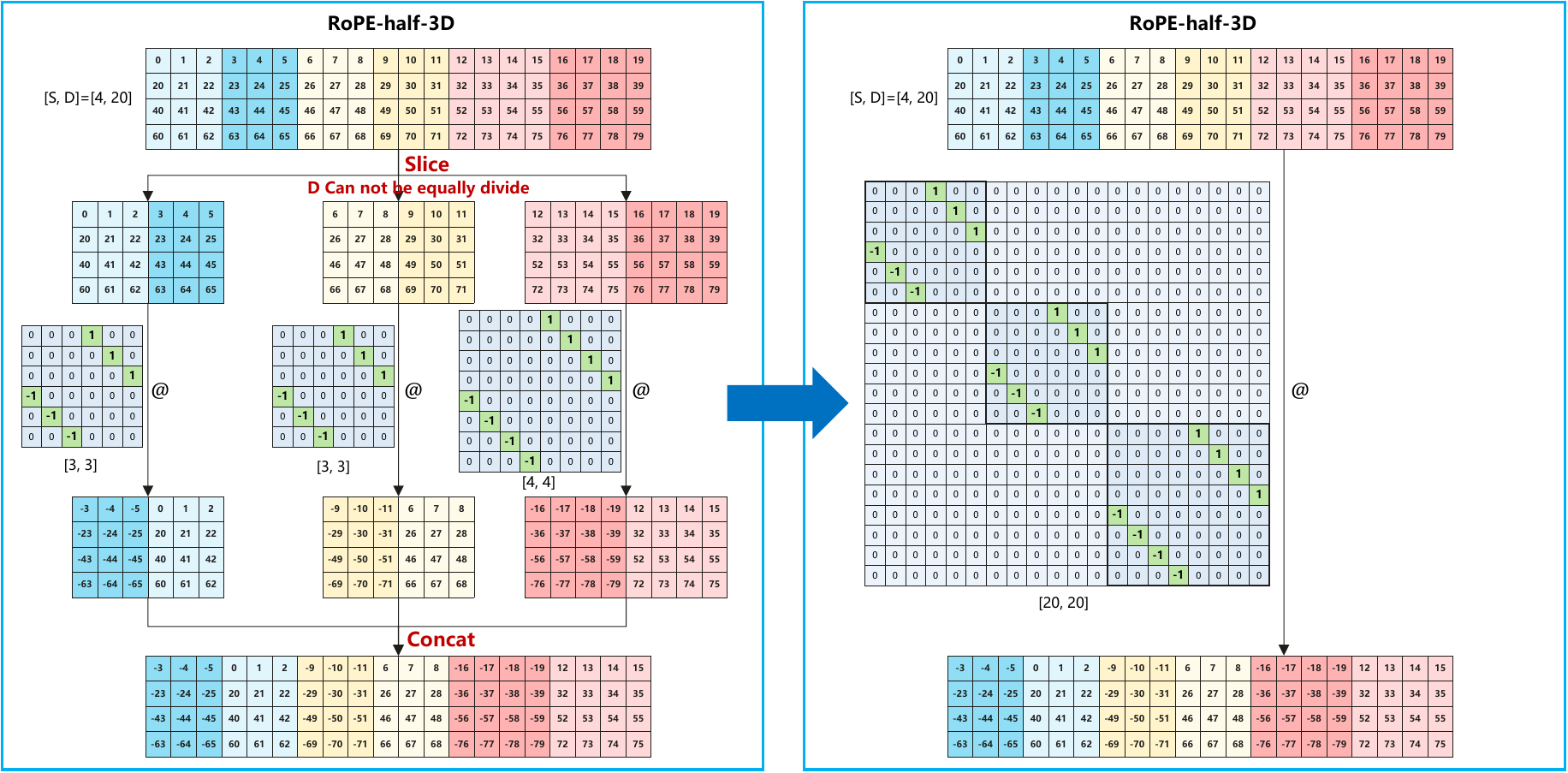}
    \caption{An example of RoPE-3D matrix.}
    \label{figure:rope_3d_matrix}
\end{figure*}

\section{Additional Experiments}